\documentclass[10pt,twocolumn,letterpaper]{article}

\usepackage{iccv}
\usepackage{times}
\usepackage{epsfig}
\usepackage{graphicx}
\usepackage{amsmath}
\usepackage{amssymb}
\usepackage{caption} 
\usepackage{colortbl} 
\usepackage{arydshln} 

\usepackage{algorithm}
\usepackage[noend]{algpseudocode}

\usepackage{multirow}
\usepackage[table,xcdraw]{xcolor}

\usepackage[breaklinks=true,bookmarks=false]{hyperref}

\iccvfinalcopy 

\def\iccvPaperID{****} 

\ificcvfinal\fi

\begin{document}

\title{3DHR-Co: A Collaborative Test-time Refinement Framework for In-the-Wild 3D Human-Body Reconstruction Task}

\author{Jonathan Samuel Lumentut$^{1}$  \hspace{45pt}  Kyoung Mu Lee$^{1,2,3}$ \\
$^{1}$IPAI, $^{2}$Dept. of ECE, ASRI, Seoul National University, $^{3}$SNU-LG AI Research Center, Korea\\
{\tt\small jslumentut@gmail.com, kyoungmu@snu.ac.kr}
}

\twocolumn[{%
\renewcommand{\arraystretch}{1.7} 
\renewcommand\twocolumn[1][]{#1}%
\ificcvfinal\thispagestyle{empty}\fi
\maketitle
\begin{center}
    \centering
    \captionsetup{type=figure}
    \includegraphics[width=\textwidth]{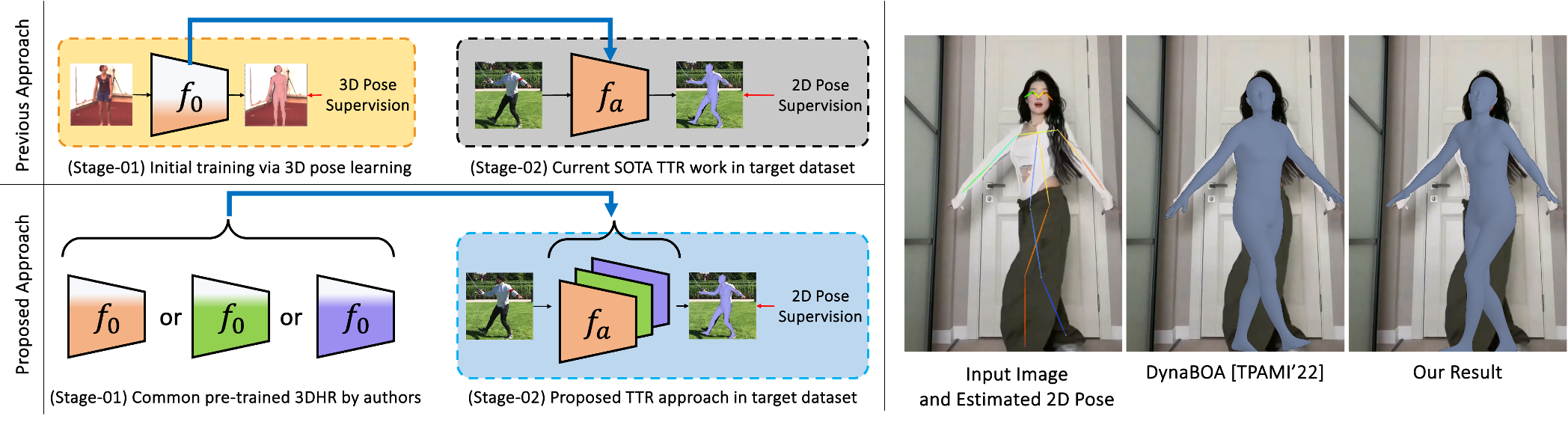}
    \captionof{figure}{We exhibit the strategy of our approach (lower branch) compared to the recent work (upper branch) in the test-time refinement strategy of 3D 3DHR on the left column. Our approach acts as a refinement tool from given pre-trained 3DHR backbones that previous works~\cite{guan_CVPR2021_bilevelbody,guan_TPAMI2022_dynaboa} was not intended to. On the right column, with given arbitrary input data, our result provided better output as it solves pose ambiguity in the final 3D 3DHR result. The recent work of DynaBOA~\cite{guan_TPAMI2022_dynaboa} utilized the ResNet-50~\cite{He_CVPR2016} backbone commonly used for 3DHR task. TTR stands for test-time refinement.}
\label{fig:fig_exhib}
\end{center}
}]

\begin{abstract}
The field of 3D human-body reconstruction (abbreviated as~\textbf{3DHR}) that utilizes parametric pose and shape representations has witnessed significant advancements in recent years. However, the application of 3DHR techniques to handle real-world, diverse scenes, known as in-the-wild data, still faces limitations. The primary challenge arises as curating accurate 3D human pose ground truth (GT) for in-the-wild scenes is still difficult to obtain due to various factors. Recent test-time refinement approaches on 3DHR leverage initial 2D off-the-shelf human keypoints information to support the lack of 3D supervision on in-the-wild data. However, we observed that additional 2D supervision alone could cause the overfitting issue on common 3DHR backbones, making the 3DHR test-time refinement task seem intractable. We answer this challenge by proposing a strategy that complements 3DHR test-time refinement work under a collaborative approach. Specifically, we initially apply a pre-adaptation approach that works by collaborating various 3DHR models in a single framework to directly improve their initial outputs. This approach is then further combined with the test-time adaptation work under specific settings that minimize the overfitting issue to further boost the 3DHR performance. The whole framework is termed as~\textbf{3DHR-Co}, and on the experiment sides, we showed that the proposed work can significantly enhance the scores of common classic 3DHR backbones up to -34 mm pose error suppression, putting them among the top list on the in-the-wild benchmark data. Such achievement shows that our approach helps unveil the true potential of the common classic 3DHR backbones. Based on these findings, we further investigate various settings on the proposed framework to better elaborate the capability of our collaborative approach in the 3DHR task.
\end{abstract}


\section{Introduction}
3DHR, which has the task of estimating 3D human pose and shape information, has gained significant attention in various applications, including AR/VR gaming, healthy-fitness tracking, and virtual cloth try-ons. 
Despite notable advancements in 3DHR works~\cite{kanazawa_CVPR18,Kolotouros_2019_SPIN,Moon_2020_I2L,fang2021reconstructing,choi_TCMR_CVPR2021}, a remaining challenge in 3DHR work remains: the availability of 3D data labels for learning is still limited.
This issue arises as acquiring the 3D labels from real-world environments~\textit{a.k.a.} in-the-wild scenario necessitates specialized equipment such as body-mounted sensors~\cite{von_3dpw_eccv2018} or multi-view cameras~\cite{mehta2018single}, which incur substantial expenses associated with data collection. 
Consequently, this scarcity of data poses a significant challenge in achieving robust 3DHR under in-the-wild scenarios.

To tackle the issue above, 3DHR works embraced the strategy of non-fully-supervised learning~\cite{kanazawa_CVPR18, kanazawa_CVPR19,Kolotouros_2019_SPIN}.
Prior to this trend, however, early works were mostly focused on the human pose estimation tasks, which the following works ~\cite{rhodin2018unsupervised,rhodin2018learning,yao2019monet} handled the few 3D label limitation by feeding the multi-view images input information.
In the upcoming years, these works by~\cite{wandt2019repnet,iqbal2020weakly,wandt2021canonpose} leveraged the weakly-supervised learning strategy that basically utilizes the non-related 2D and 3D label-based data together for training their human pose estimation network.


As the above's human pose estimation works showed considerable performances, pioneer works~\cite{kanazawa_CVPR18, kanazawa_CVPR19,Kolotouros_2019_SPIN} are proposed to directly infer the human pose and shape outputs from the image or video input via parametric function (e.g. SMPL~\cite{loper_ACM15}, MANO~\cite{romero2022embodied_siggraph2017}).
Those parametric-based works are also known for their straightforward benefit during learning as they can utilize the estimated 3D human pose parameters that are projected to 2D representations, to be supervised by the richly-available 2D ground truth dataset (such as the dataset of MSCOCO~\cite{lin2014mscoco}).
These works were then evolved into the temporal strategy (\textit{e.g.}:~\cite{kocabas_vibe_CVPR2020,luo_meva_ACCV2020,choi_TCMR_CVPR2021,wei_CVPR2022_mpsnet}) that basically utilized features of neighboring frames of video input to subdue the lack of feature information in particular frame or sequences due to \textit{unseen} body parts~\cite{kocabas_pare_ICCV2021}.

As those recent 3DHR works~\cite{kanazawa_CVPR18, kanazawa_CVPR19,Kolotouros_2019_SPIN} showed considerable ability in predicting human pose and shape outputs by including 2D data supervision, recent approaches, such as BOA~\cite{guan_CVPR2021_bilevelbody} and DynaBOA~\cite{guan_TPAMI2022_dynaboa} utilized the 2D detected pose information provided from the off-the-shelf detector to firmly guide the 3DHR model during inference-time learning or known as test-time adaptation.
Such strategies pave the novel way to achieve top performance in the in-the-wild benchmark data (such as 3DPW~\cite{von_3dpw_eccv2018}) without much modification on the network architecture level.
Their works, however, are only focused on the task of domain adaptation.
As shown in the left column's upper branch of Figure~\ref{fig:fig_exhib}, they (BOA or DynaBOA) require an initial training step with specific dataset ~(\textit{e.g.}: Human3.6M~\cite{ionescu_TPAMI14_human36m}) that is limited in terms of variety but provides 3D ground truth, to be later adapted on test time with both Human3.6M~\cite{ionescu_TPAMI14_human36m} (as~\textit{Source} domain) and in-the-wild data (\textit{e.g.} 3DPW~\cite{von_3dpw_eccv2018} as~\textit{Target} domain).

\begin{figure}[t]
\begin{center}
        \includegraphics[width=0.45\textwidth]{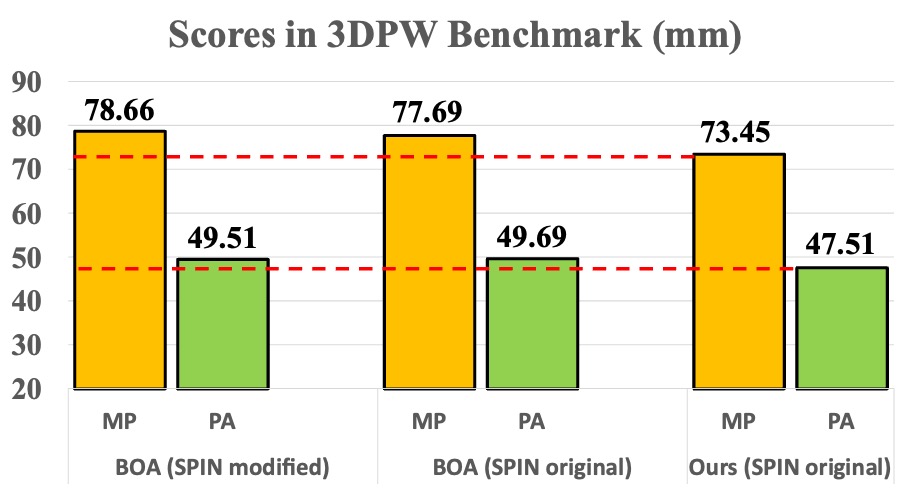} 
\end{center}
\caption{The error scores of BOA~\cite{guan_CVPR2021_bilevelbody} with original SPIN~\cite{Kolotouros_2019_SPIN} backbone pre-trained with various datasets (middle) and with modified SPIN backbone pre-trained with single dataset (left) that is on-par. We tackled this issue by performing a refinement strategy that pushed the error score lower (right).}
\label{fig:fig02_motivation}
\end{figure}

Based on the matter above, our observation in Figure~\ref{fig:fig02_motivation} showed that the SPIN's~\cite{Kolotouros_2019_SPIN} backbone's performance (middle bars of Figure~\ref{fig:fig02_motivation}) that are directly plugged onto BOA's framework (\textit{BOA-plugged}), was only showing on-par results with the original BOA's work that has its own modified backbone version (left-bars).
This is detrimental as \textit{BOA-plugged}, pre-trained with various datasets, should have shown better performance than the original BOA, which was solely trained with Human3.6M~\cite{ionescu_TPAMI14_human36m}.
This issue is caused by the overfitting problem on \textit{BOA-plugged}, as it directly adapts the whole video data in the target domain, meaning that any video frame sequences unrelated to the initially fetched frames are insufficient for adaptation.

We tackled the issue above with a straightforward objective: proposing a refinement framework that can plug the original pre-trained 3DHR model $f_0$ and directly show better refined 3D body results in test time than the initial version via its adapted version $f_a$.
The general idea of this setup is displayed in the left column's lower branch part of Figure~\ref{fig:fig_exhib}.
The knowledge of each backbone is represented with different colors, and our test-time refinement strategy is meant to increase each's knowledge to predict better 3DHR outputs.
Following the idea of BOA (or DynaBOA) that utilizes off-the-shelf 2D pose information during adaptation, our work was able to show better visual results (right column of Figure~\ref{fig:fig_exhib}) than the in-the-wild 3DPW~\cite{von_3dpw_eccv2018}'s recent adaptive-based method (DynaBOA~\cite{guan_TPAMI2022_dynaboa}).

Our strategy, termed as~\textit{3DHR-Co}, is a test-time refinement approach that utilizes currently available 3DHR methods to work collaboratively in distilling the information from one model (as a teacher) unto another (as a learner).
This is made possible by treating the learner side of the 3DHR model with the perturbed version while the teacher side with the non-perturbed version of input data during test time.
The 3DHR outputs difference between perturbed and non-perturbed versions provide implicit cues for learning during inference time, making it similar to the strategy self-supervised zero-shot learning~\cite{shocherzero_cvpr2018}.
In the following sections of this manuscript, we provide further discussion related to our works and the detailed strategies for better understanding.
To summarize, our contributions above are written as follows:
\begin{itemize}
\item The proposal of a collaborative-based test-time refinement strategy for 3DHR task (3DHR-Co) that can refine 3DHR models on the go.
\item Top-level performance achievement on the in-the-wild benchmark data, obtained by only using the classic common 3DHR backbones run via our collaborative-based test-time refinement framework. 
\item A study that provides discussions and recommendations regarding the optimal solution for running 3DHR test-time refinement under collaborative strategy. 
\end{itemize}



\section{Related Works}
We describe several prior works that are relevant to our main studies. 
They are classified and described in the following subsections:
\subsection{3D Human-body Reconstruction}
Recently, 3D human-body reconstruction has gained popularity in the computer vision research community.
Early works focused on estimating the pose of humans in a stick-man representation, mostly known as the task of human pose estimation~\cite{rhodin2018unsupervised,rhodin2018learning,yao2019monet,wandt2019repnet,iqbal2020weakly,wandt2021canonpose}.
In the past few years, parametric-based approaches have been proposed to simplify the prediction task of 3DHR by estimating the human pose and shape outputs directly.
Among these parametric-based approaches, one particular work (namely skinned multi-person linear model or SMPL~\cite{loper_ACM15}) was proposed to represent the non-clothed human mesh form with those parameters.
Early work termed as Human Mesh Recovery (HMR)~\cite{kanazawa_CVPR18} utilized deep learning architecture~\cite{He_CVPR2016} to provide features for regressing the SMPL parameters and camera outputs to render the final human mesh output.
The following works then utilized a similar strategy but with several modifications on the backbone levels to support better 3DHR performance from single input image~\cite{Xu_ECCV2020_LRBody,choi2020pose2mesh}. 
Realizing that image-based works can be extended to the temporal domain (video-based), various works are also proposed to tackle the limitation of single image input-based 3DHR works~\cite{kanazawa_CVPR19,luo_meva_ACCV2020,choi_TCMR_CVPR2021,wei_CVPR2022_mpsnet}.
These approaches are naturally better as temporal strategy provides features from neighboring frames that help recognize certain human poses that might be undetected in a particular frame.
Just a while ago, recent works~\cite{hanbyul_expressivemodel_CVPR2018,pavlakos_smplx_cvpr2019,xu_cvpr2020_ghuml} involved particular expressive parameters such as hand and face details.
Moreover, the current developments~\cite{Shunsuke_2019_PIFu,saito_pifuhd_cvpr2020,he_archplus_iccv2021,alldieck_photorealistic_cloth_cvpr2022} also included clothing details information which is pleasing to look on.

\subsection{Test-time Learning}

Test-time learning, also known as test-time training, is an approach that performs learning on the deep network model during inference time that aims to increase the knowledge of the model itself to be more robust in solving certain task~\cite{sun_catastroph_forget_ICML2020}.
Other non-3DHR works~\cite{shocherzero_cvpr2018,ehret_frame2frame_cvpr2019, lee_restore_pseudo_CVPR2021} have experimented with test-time learning to boost their performances.
One interesting work, zero-shot image restoration~\cite{shocherzero_cvpr2018}, showed that test-time learning is possible from an untrained model.
This strategy is further improved by these recent works that mainly utilized the meta-learning approach to firstly train the untrained model and then do the test-time learning procedure for fast adaptation purpose~\cite{KimFast_ECCV2020,SohMeta_CVPR2020,Lee_DynaVidSR_WACV21}.
In recent times, however, the paradigm of test-time learning has intersected with the idea of domain adaptation, which aims to adapt a model learned from a specific domain to be robust on certain target domains.
This idea later evolved into domain generalization, where the model is adapted from certain to generalize well to out-of-distribution various target data.
In relation to the 3DHR works recent studies of BOA~\cite{guan_CVPR2021_bilevelbody} and DynaBOA~\cite{guan_TPAMI2022_dynaboa} implemented the idea of test-time training for domain adaptation.
DynaBOA~\cite{guan_TPAMI2022_dynaboa} utilized additional information retrieval and dynamic adaptation to surpass its predecessor (BOA~\cite{guan_CVPR2021_bilevelbody}).
These prior works differ from us in terms of motivation, as they focus on the task of solving domain shift issues via test-time adaptation.

\section{Method}
\subsection{Preliminary}

\begin{figure}[t]
\begin{center}
        \includegraphics[width=0.48\textwidth]{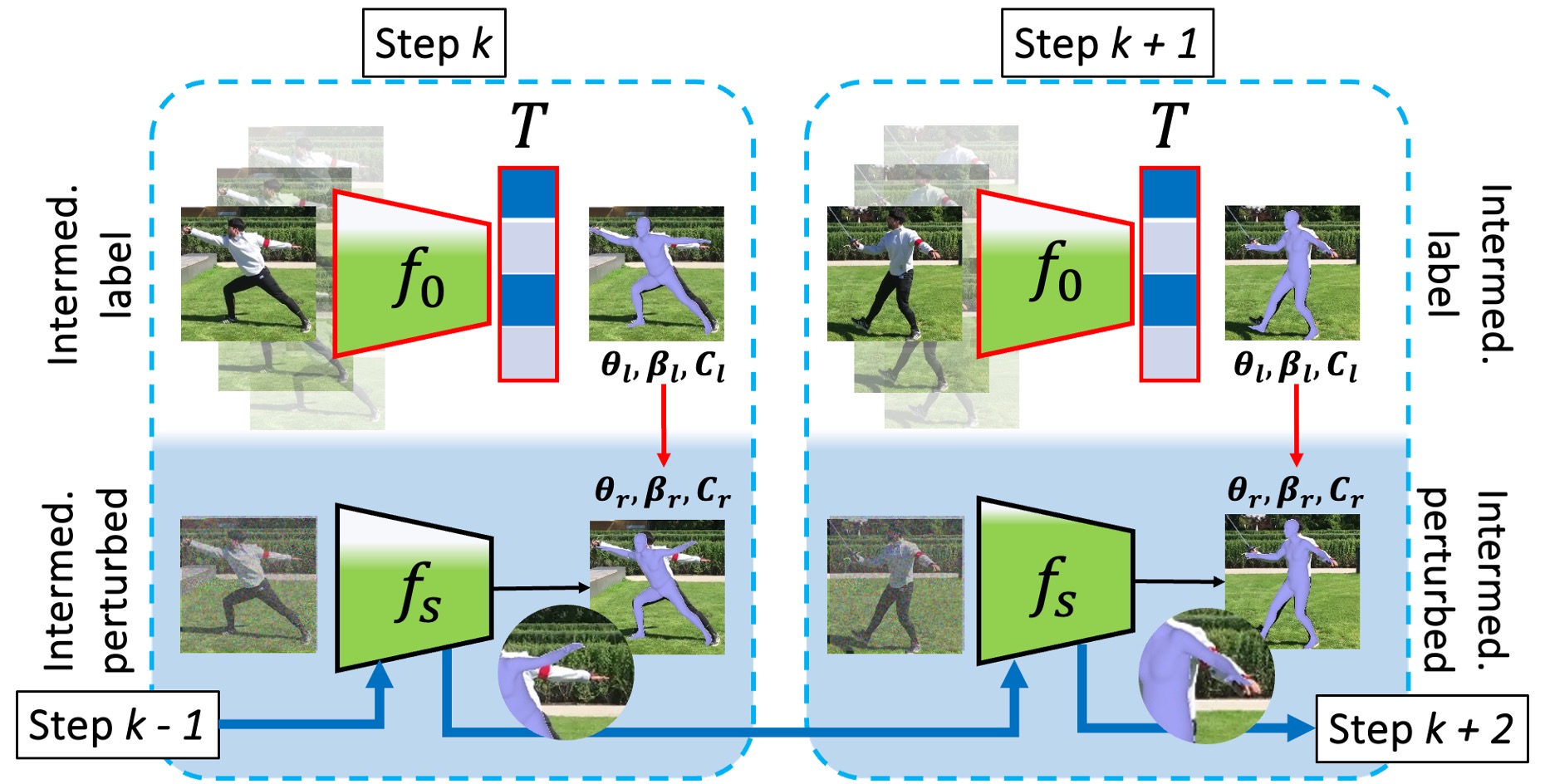}
\end{center}
\caption{Our straightforward yet intuitive test-time refinement framework. The framework above denotes our pre-adaptations strategy (\textbf{Line 6-13} of Algorithm~\ref{alg:algo_hmrco_ttr}), which aims to provide pre-adapted weight ($f_{s}$). The form of $f_{s}$ is already within our test-time refinement framework; however, it is best to furtherly refine it using bilevel adaptation~\cite{guan_CVPR2021_bilevelbody,guan_TPAMI2022_dynaboa}.}
\label{fig:fig04_main_strategy}
\end{figure}

In this passage, we describe the general idea of our approach.
In contrast to prior works~\cite{guan_CVPR2021_bilevelbody,guan_TPAMI2022_dynaboa}, ours is focused on the task of refinement strategy of the pre-trained-available 3DHR models. 
The objective above led us to the strategy of collaborative learning, where two different 3DHR models are placed directly into our refinement framework.
As shown in Figure~\ref{fig:fig04_main_strategy}, our framework provided 2 specific branches (white and blue regions) with one model ($f_0$ placed on the white-colored upper-branch region) acting as the teacher and the other one ($f_s$ placed on the blue-colored lower-branch region) acts as learner.
The knowledge of the teacher model with no perturbance case is transferred to the learner model that is perturbed.
The perturbed version is provided with noise addition on the image level that acts as a partial occlusion to the human body.
By teaching the learner model with the knowledge of the non-perturbed version, it is natural that the learner model gains more improvement during refinement (visualized by the increased green-colored content of $f_s$ in Figure~\ref{fig:fig04_main_strategy}).
Based on the idea above, we provide a detailed elaboration of our test-time refinement mechanism for the 3DHR task in the following.
 
For simplicity purposes, we make use of the common 3DHR methods that utilize the SMPL parametric model~\cite{loper_ACM15}. 
The SMPL model is constructed by body pose parameter $\theta$ and the shape parameter $\beta$. 
In addition, besides predicting the SMPL parameters, recent common 3DHR methods are also tasked to estimate the camera parameters, $c$.
These SMPL parameters ($\theta$, $\beta$) can be used to generate corresponding human mesh representation, $\mathcal{M}$, along with the 3D keypoints representation $J$ via the mesh-to-3D-skeleton mappings provided in SMPL function.
The camera parameter $c$ can be used to perform a weak-perspective projection of $J$ from 3D to 2D space.
The above nomenclatures are used in the following elaboration.

\begin{algorithm*}[t]
\caption{3DHR-Co Test-time Refinement Algorithm}
\label{alg:algo_hmrco_ttr}
\begin{algorithmic}[1]
\State \textbf{Input}: Images from benchmark dataset ($X$)
\State \textbf{Require}: Pre-trained 3DHR ($f_0$) and 2D pose guide ($G$)
\State \textbf{Output}: Refined body ($\hat{\theta}$, $\hat{\beta}$) and camera ($\hat{C}$) parameters
\State $f_{s}$ = deepcopy($f_0$) \Comment{Create learner version of $f_0$} 
\State $B$ = load($X$) \Comment{Create batch collections $B$ from $X$}
\While{step $k < K$ and $K \in$ epoch}

    \State $B_{i}$ = iter\_fetch($B$) \Comment{Fetch each data from loader}
    \State $B_{r}$ = corr($B_{i}$) \Comment{Generate corrupted input}
    \State $V_{l}$ = $f_0^{'}([B_{i-j}, ...,B_i, ..., B_{i+j}])$ \Comment{Extract temporal features}
    \State ($\theta_{l}$, $\beta_{l}$, $C_{l}$) = $T(V_{l})$ \Comment{Extract intermediate labels outputs}
    \State ($\theta_{r}$, $\beta_{r}$, $C_{r}$) = $f_s$($B_{r}$) \Comment{Extract intermediate perturbed outputs}
    \State $\mathcal{L} = $ loss($\theta_{l}$, $\beta_{l}$, $C_{l}$, $\theta_{r}$, $\beta_{r}$, $C_{r}$, $G$) by using Eq.~(\ref{eq:eq_01_loss_preadaptation})
    \State Update learner: $f_{s}$ $\leftarrow$ ADAM($f_{s}$, $\nabla_f\mathcal{L}$ )
    
\EndWhile 
\State $P$ = [ ] \Comment{Create buffer info for knowing latest sequence}
\While{batch $B$ is available}
    \If{$B$.name != $P$.name}
        $f_a$ = $f_s$ \Comment{Regenerate new $f_a$}
    \EndIf

    \State ($\hat{\theta}$, $\hat{\beta}$, $\hat{C}$, $f_{a}$) = Bilevel($f_a$, $B$, $G$) \Comment{Extract \textbf{refined} outputs}
    
    \State $P$ = $B$ \Comment{Update buffer $P$ with the latest iteration data}
\EndWhile
\end{algorithmic}
\end{algorithm*}

\subsection{3DHR-Co test-time refinement algorithm} 
\subsubsection{Motivation}
Our test-time refinement framework is explicitly defined in an algorithm form, and its representation is outlined in the Algorithm~\ref{alg:algo_hmrco_ttr}.
As illustrated in Figure~\ref{fig:fig04_main_strategy} earlier, our method is designed to directly refine the pre-trained 3DHR model. 
To achieve this, we employ two refinement strategies: (i) \textit{\textbf{the pre-adaptation stage}} (\textbf{Line 6-13} in Algorithm~\ref{alg:algo_hmrco_ttr}), and (ii) \textit{\textbf{the full scope}} of our 3DHR-Co test-time refinement, which includes further refinement through our regeneration-based bilevel adaptation (\textbf{Line 15-18}) in Algorithm~\ref{alg:algo_hmrco_ttr}). 
The purpose of our pre-adaptation strategy (depicted in Figure~\ref{fig:fig04_main_strategy}) is to provide a ready-to-adapt weight that avoids overfitting issues when the bilevel adaptation algorithm is plugged directly with the 3DHR backbone. 
As bilevel strategy works in a sequential manner, the backbone has the tendency to preserve knowledge on certain sequential frames, leaving the remaining incoming streams to be sub-optimal during adaptation.
Our strategy tackled this issue by introducing the sampled target data during the pre-adaptation stage.
The effect is crucial, and as demonstrated in Figure~\ref{fig:fig02_motivation}, our method (right-bars) with BOA function and the pre-adapted SPIN weight can achieve better results than the BOA function with the pre-trained SPIN weight (middle-bars).
With this motivation, the whole 3DHR-Co test-time refinement algorithm is constructed to suit the plugged pre-trained model.

\subsubsection{Pre-adaptation Stage}
\label{sec:sec_preadapt}
In this passage, we provide the technical elaboration on running the pre-adaptation strategy.
The pre-adaptation strategy is visually depicted in Figure~\ref{fig:fig04_main_strategy} and pseudocode-wise, it is written in the \textbf{Line 6-13} in Algorithm~\ref{alg:algo_hmrco_ttr}.
To perform pre-adaptation, our work is supported with 2 intermediate output data, namely: test-time \textit{\textbf{intermediate label}} and \textit{\textbf{intermediate perturbed}} data.
The intermediate label data is generated by the non-perturbed version of test input data, while the perturbed version acts as the intermediate perturbed data for the learner module.
This approach came with 2 benefits: (a) arbitrary in-the-wild input data can be used directly to extract the perturbed data for refinement purposes, and (b) various models without modification can be learned on the go during test-time.
The remaining task is then focused on the strategy of creating a reliable test-time refinement framework.

At the algorithm level, our pre-adaptation approach first fetches the current batch data $B_i$ and its neighboring frames ($B_{i-j}, ... , B_{i+j}$) for extracting the temporal features $V_l$.
This procedure is done to extract the intermediate label data ($ \theta_{l}, \beta_{l}, C_{l}$ in \textbf{Line 10}) extractor, and in this work, the common temporal~\cite{choi_TCMR_CVPR2021,wei_CVPR2022_mpsnet} 3DHR works are directly utilized to fulfill such task.
Meanwhile, the intermediate perturbed data ($ \theta_{r}, \beta_{r}, C_{r}$ in \textbf{Line 11}) version is obtained from the predicted body and camera parameters from test input data $B_r$ corrupted via Gaussian noise (\textbf{Line 8}).

This strategy is further learned via MSE loss functionality (\textbf{Line 12}): 
\begin{equation}
\label{eq:eq_01_loss_preadaptation}
\begin{array}{l}
\mathcal{L} = \lambda_{1} \| \theta_{l} - \theta_{r}\|^2 + \lambda_{2} \| \beta_{l} - \beta_{r}\|^2 \\
\hspace{20mm} + \lambda_{3} \| C_{l} - C_{r}\|^2 + \lambda_{4} \| G - C_{r}(J_r)\|^2,      
\end{array}
\end{equation}
and its visual representation is denoted by the red arrow of Figure~\ref{fig:fig04_main_strategy} where each of the output parameters is used for supervision.
$J_r$ is the 3D keypoint information extracted from the SMPL output parameters of the perturbed version that are projected into 2D via $C_{r}$.
The approach above simulates continual knowledge improvement (step-by-step wise) as depicted by the increased green-colored content in the lower-branch scope of the student network ($f_s$) in Figure~\ref{fig:fig04_main_strategy}.
Note that the batch of data in each step is sampled randomly. 
Thus on the step of $k+1$, data can be unrelated to data in the previous step $k$ as shown in Figure~\ref{fig:fig04_main_strategy}.
Once the pre-adaptation stage is finished, $f_{a}$ is transferred to the next scope (\textbf{Line 15-18}) to perform bilevel test-time refinement.

\subsubsection{Regeneration-based bilevel test-time refinement}
The full-scope of the 3DHR-Co refinement framework includes the initial pre-adaptation strategy above and the following regeneration-based bilevel refinement approach (\textbf{Line 15-18}).
The objective of the latter's strategy is to refine the pre-adapted 3DHR backbone's weights while also avoiding the overfitting issue.
To run the regeneration strategy, we employ a straightforward weight refreshment mechanism (\textbf{Line 16}) that runs under frame sequence manner. 
The data buffer $P$ is utilized to store the latest adapted weight that learns from previous stream sequence data (\textbf{Line 18}).
For the adaptation, the bilevel function (\textbf{Line 17}) is applied, and in this step, $f_a$ is always updated along the batch input.
The bilevel function acts as a refinement tool that can adapt the pre-adapted $f_a$ 3DHR model in test-time, and we refer to the original implementation of the respective authors~\cite{guan_CVPR2021_bilevelbody,guan_TPAMI2022_dynaboa}.

The regeneration strategy is proposed to avoid the overfitting issue mentioned earlier.
The procedure is straightforward (\textbf{Line 16}) as it re-initiates the model with the weight obtained from the pre-adaptation stage (\textbf{Line 6-13}) when a new sequence is fetched.
Doing so helps the model to directly re-learn from the pre-adapted version $f_s$, which already has the knowledge of tested data, rather than the pre-trained version $f_0$.
This step is important as the $f_s$ version already secured initial improvement capability without being overfitted to a particular frame or sequence on test data.
These capabilities are further measured in the following \textit{Experiment} section.

\begin{figure}[t]
\begin{center}
        \includegraphics[width=0.47\textwidth]{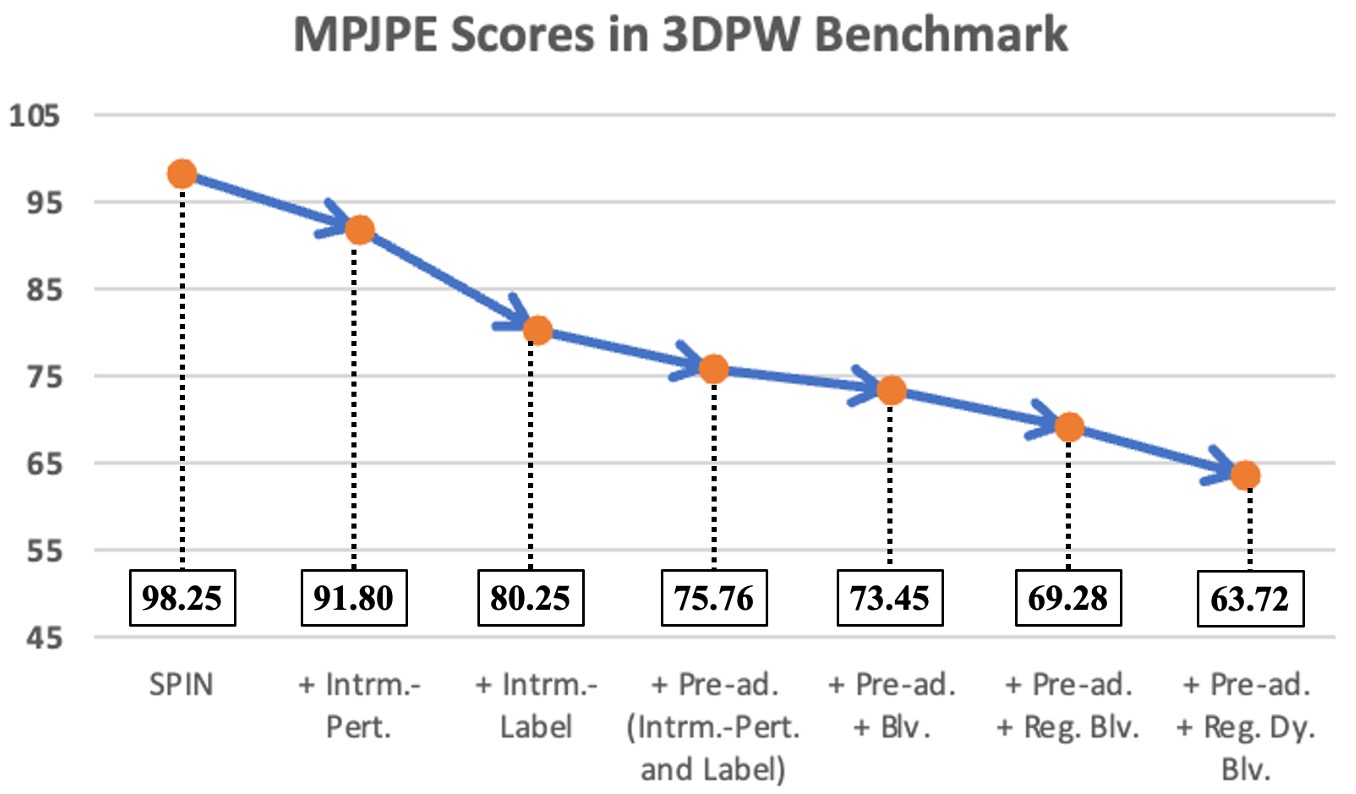}
\end{center}
\caption{Improvement of the classic SPIN~\cite{Kolotouros_2019_SPIN} backbone in stage-by-stage manner. By accumulating the additional functions proposed in this work, classic SPIN backbone can achieve state-of-the-art result in in-the-wild benchmark data (3DPW~\cite{von_3dpw_eccv2018}). Legends: \textbf{Intrm.}=intermediate, \textbf{pert.}=perturbed, \textbf{pre-ad.}=pre-adaptation, \textbf{reg.}=regeneration, \textbf{dy.}=dynamic, and \textbf{blv.}=bilevel.}
\label{fig:fig03_improvementtosota}
\end{figure}

\begin{figure*}[t]
\begin{center}
        \includegraphics[width=0.85\textwidth]{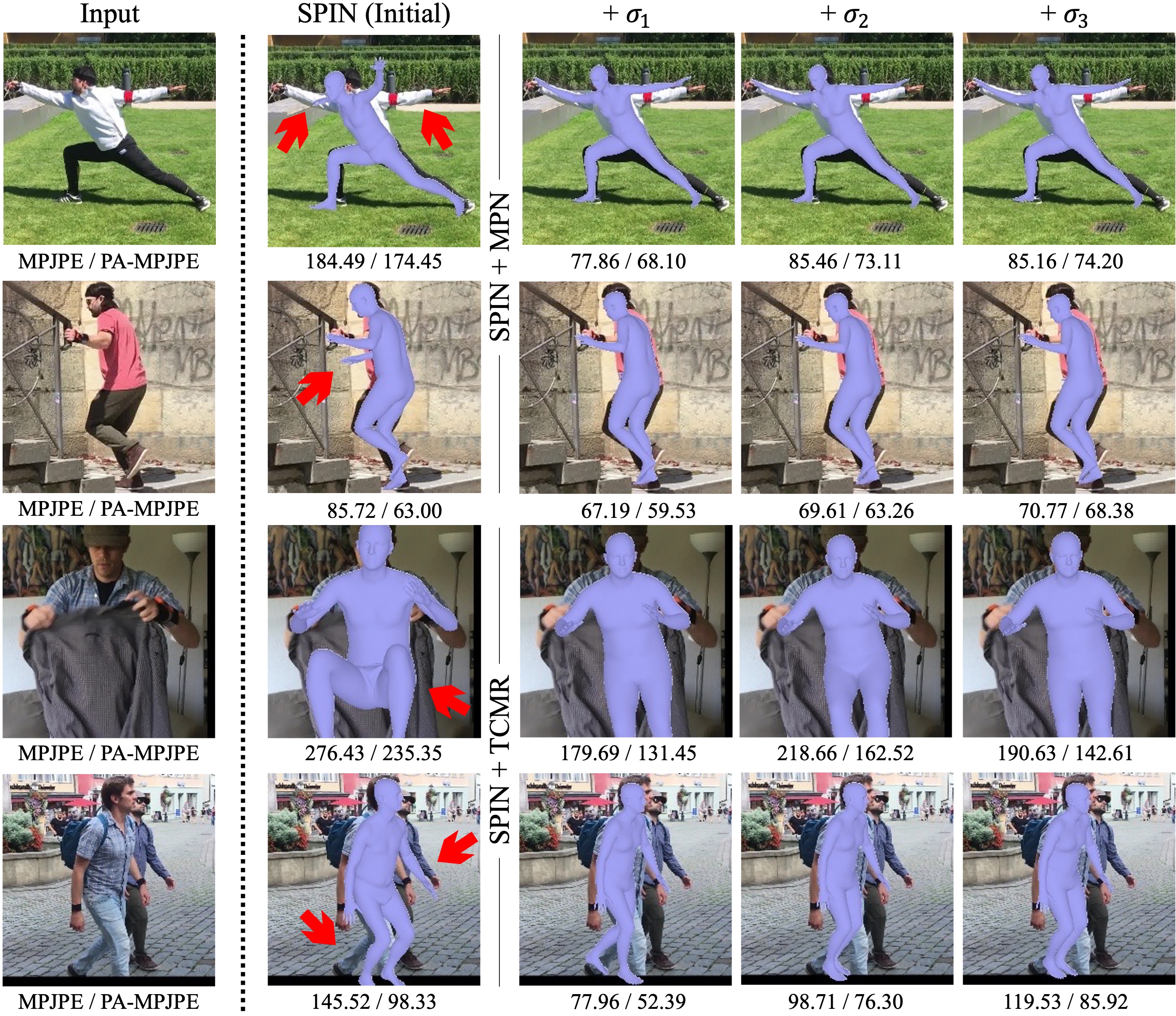}
\end{center}
\caption{Qualitative results of \textbf{SPIN} method that are adapted using our pre-adaptation strategy. Significant pose changes are highlighted with red arrow marks. MPN and TCMR act as the intermediate label extractor during pre-adaptation.}
\label{fig:fig05_SPIN_abla}
\end{figure*}

\begin{figure*}[t]
\begin{center}
        \includegraphics[width=0.85\textwidth]{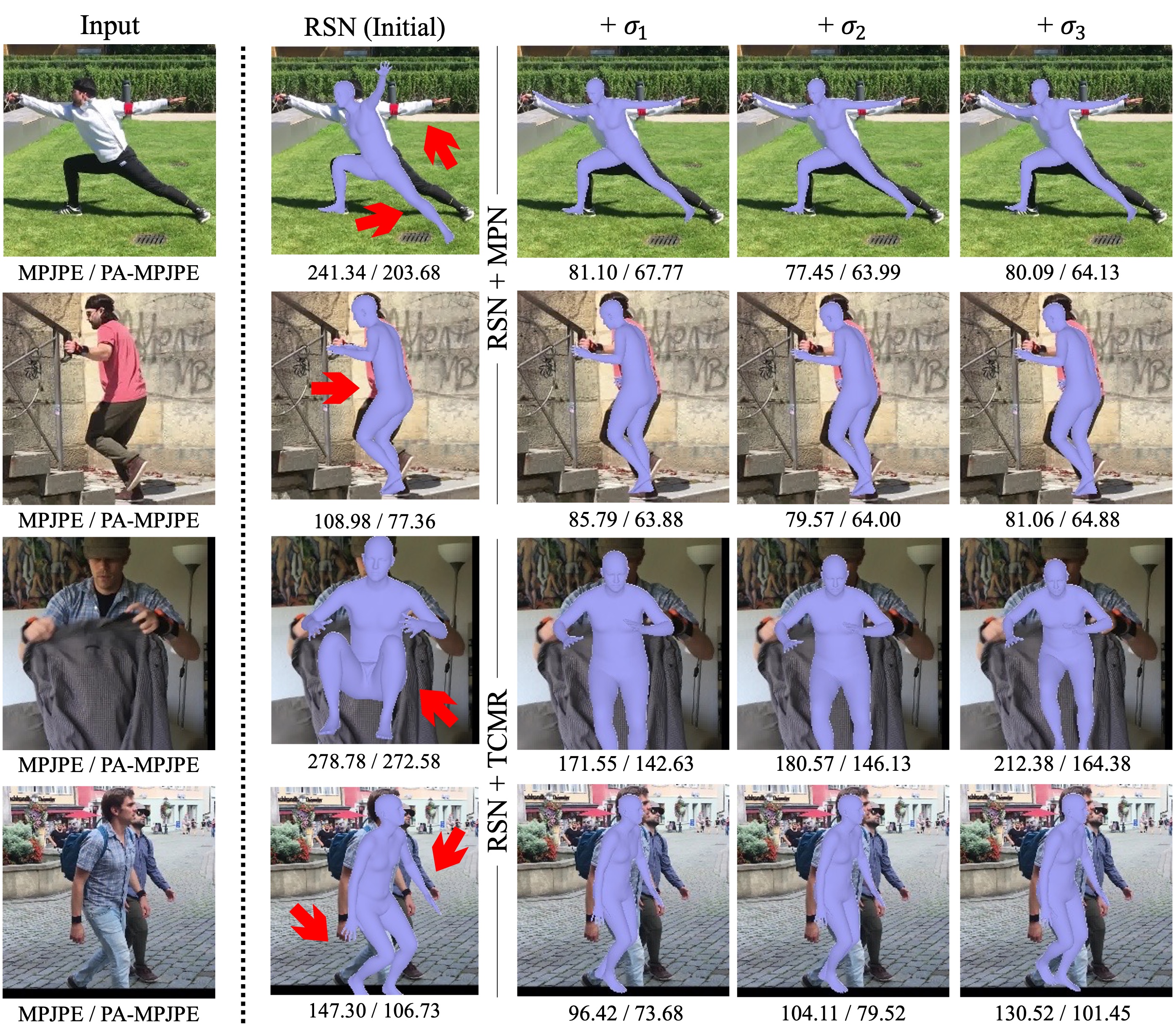}
\end{center}
\caption{Qualitative results of \textbf{RSN} method that are adapted using our pre-adaptation strategy. Significant pose changes are highlighted with red arrow marks. MPN and TCMR act as the intermediate label extractor during pre-adaptation.}
\label{fig:fig05_RSN_abla}
\end{figure*}

\begin{table*}[t]
\caption{Quantitative scores by performing ablation study on the number of sampled data (percentage) and noise information in our~\textbf{pre-adaptation strategy}. In the following, \textbf{SPIN}~\cite{Kolotouros_2019_SPIN} scores improvement are shown. MPN~\cite{wei_CVPR2022_mpsnet} and TCMR~\cite{choi_TCMR_CVPR2021} are selected as the intermediate label data generator. Our algorithm allows all methods to work collaboratively in the test-time refinement mechanism. {\color[HTML]{F56B00}Orange} mark defines error gap $<$ -20 mm.} 
\label{table:tab_SPIN_abla}
\resizebox{\linewidth}{!}{%
\begin{tabular}{cc|c|cccccccccccc}
\hline
\multicolumn{2}{c|}{}                                             &                           & \multicolumn{3}{c}{ep-1 (0.07\%)}                                              & \multicolumn{3}{c}{ep-201 (14\%)}                                                                                & \multicolumn{3}{c}{ep-401 (28\%)}                                                                               & \multicolumn{3}{c}{ep-601 (42\%)}                                                          \\ \cline{4-15} 
\multicolumn{2}{c|}{\multirow{-2}{*}{Intrm. {\color[HTML]{009900}-Label} \textbackslash~{\color[HTML]{0000ff}-Pert.}}} & \multirow{-2}{*}{SPIN} & \multicolumn{1}{c|}{{\color[HTML]{0000ff}$\sigma_1$}} & \multicolumn{1}{c|}{{\color[HTML]{0000ff}$\sigma_2$}} & \multicolumn{1}{c|}{{\color[HTML]{0000ff}$\sigma_3$}}    & \multicolumn{1}{c|}{{\color[HTML]{0000ff}$\sigma_1$}}       & \multicolumn{1}{c|}{{\color[HTML]{0000ff}$\sigma_2$}}       & \multicolumn{1}{c|}{{\color[HTML]{0000ff}$\sigma_3$}}                            & \multicolumn{1}{c|}{{\color[HTML]{0000ff}$\sigma_1$}}       & \multicolumn{1}{c|}{{\color[HTML]{0000ff}$\sigma_2$}}       & \multicolumn{1}{c|}{{\color[HTML]{0000ff}$\sigma_3$}}                            & \multicolumn{1}{c|}{{\color[HTML]{0000ff}$\sigma_1$}}       & \multicolumn{1}{c|}{{\color[HTML]{0000ff}$\sigma_2$}}       & {{\color[HTML]{0000ff}$\sigma_3$}}                            \\ \hline
\multicolumn{2}{c|}{Initial}                                      & 98.25                     & -                       & -                       & \multicolumn{1}{c|}{-}     & -                             & -                             & \multicolumn{1}{c|}{-}                             & -                             & -                             & \multicolumn{1}{c|}{-}                             & -                             & -                             & -                             \\ \hline
\multicolumn{1}{c|}{}                             & MPJPE         & -                         & 93.26                   & 94.08                   & \multicolumn{1}{c|}{93.26} & 80.77                         & 82.93                         & \multicolumn{1}{c|}{84.36}                         & 78.83                         & 79.70                         & \multicolumn{1}{c|}{82.44}                         & 76.44                         & 77.42                         & 79.92                         \\
\multicolumn{1}{c|}{\multirow{-2}{*}{{\color[HTML]{009900}MPN}}}        & Gap      & -                         & -4.99                   & -4.17                   & \multicolumn{1}{c|}{-4.99} & {\color[HTML]{000000} -17.48} & {\color[HTML]{000000} -15.32} & \multicolumn{1}{c|}{{\color[HTML]{000000} -13.89}} & {\color[HTML]{000000} -19.42} & {\color[HTML]{000000} -18.55} & \multicolumn{1}{c|}{{\color[HTML]{000000} -15.81}} & {\color[HTML]{F56B00} -21.81} & {\color[HTML]{F56B00} -20.83} & {\color[HTML]{000000} -18.33} \\ \hline
\multicolumn{1}{c|}{}                             & MPJPE         & -                         & 92.90                   & 93.95                   & \multicolumn{1}{c|}{94.47} & 79.86                         & 81.22                         & \multicolumn{1}{c|}{83.72}                         & 77.72                         & 82.56                         & \multicolumn{1}{c|}{82.07}                         & 75.76                         & 77.10                         & 79.18                         \\
\multicolumn{1}{c|}{\multirow{-2}{*}{{\color[HTML]{009900}TCMR}}}       & Gap      & -                         & -5.35                   & -4.31                   & \multicolumn{1}{c|}{-3.79} & {\color[HTML]{000000} -18.39} & {\color[HTML]{000000} -17.03} & \multicolumn{1}{c|}{{\color[HTML]{000000} -14.53}} & {\color[HTML]{F56B00} -20.53} & {\color[HTML]{000000} -15.70} & \multicolumn{1}{c|}{{\color[HTML]{000000} -16.18}} & {\color[HTML]{F56B00} -22.49} & {\color[HTML]{F56B00} -21.15} & {\color[HTML]{000000} -19.07} \\ \hline
\end{tabular}
} 
\end{table*}

\begin{table*}[t]
\caption{Quantitative scores by performing ablation study on the number of sampled data (percentage) and noise information in our~\textbf{pre-adaptation strategy}. In the following, \textbf{RSN}~\cite{Xu_ECCV2020_LRBody} scores improvement are shown. MPN~\cite{wei_CVPR2022_mpsnet} and TCMR~\cite{choi_TCMR_CVPR2021} are selected as the intermediate label data meshes generator. Our algorithm allows all methods to work collaboratively in the test-time refinement mechanism. {\color[HTML]{F56B00}Orange} mark defines error gap $<$ -20 mm.}  
\label{table:tab_RSN_abla}
\resizebox{\linewidth}{!}{%
\begin{tabular}{cl|c|cccccccccccc}
\hline
\multicolumn{2}{c|}{}                                             &                          & \multicolumn{3}{c}{ep-1 (0.07\%)}                                                                                  & \multicolumn{3}{c}{ep-201 (14\%)}                                                                                & \multicolumn{3}{c}{ep-401 (28\%)}                                                                               & \multicolumn{3}{c}{ep-601 (42\%)}                                                          \\ \cline{4-15} 
\multicolumn{2}{c|}{\multirow{-2}{*}{Intrm. {\color[HTML]{009900}Label} \textbackslash~{\color[HTML]{0000ff}Pert.}}} & \multirow{-2}{*}{RSN} & \multicolumn{1}{c|}{{\color[HTML]{0000ff}$\sigma_1$}}       & \multicolumn{1}{c|}{{\color[HTML]{0000ff}$\sigma_2$}}       & \multicolumn{1}{c|}{{\color[HTML]{0000ff}$\sigma_3$}}                            & \multicolumn{1}{c|}{{\color[HTML]{0000ff}$\sigma_1$}}       & \multicolumn{1}{c|}{{\color[HTML]{0000ff}$\sigma_2$}}       & \multicolumn{1}{c|}{{\color[HTML]{0000ff}$\sigma_3$}}                            & \multicolumn{1}{c|}{{\color[HTML]{0000ff}$\sigma_1$}}       & \multicolumn{1}{c|}{{\color[HTML]{0000ff}$\sigma_2$}}       & \multicolumn{1}{c|}{{\color[HTML]{0000ff}$\sigma_3$}}                            & \multicolumn{1}{c|}{{\color[HTML]{0000ff}$\sigma_1$}}       & \multicolumn{1}{c|}{{\color[HTML]{0000ff}$\sigma_2$}}       & {\color[HTML]{0000ff}$\sigma_3$}                            \\ \hline
\multicolumn{2}{c|}{Initial}                                       & 98.51                    & -                             & -                             & \multicolumn{1}{c|}{-}                             & -                             & -                             & \multicolumn{1}{c|}{-}                             & -                             & -                             & \multicolumn{1}{c|}{-}                             & -                             & -                             & -                             \\ \hline
\multicolumn{1}{c|}{}                             & MPJPE            & -                        & 89.14                         & 86.52                         & \multicolumn{1}{c|}{83.58}                         & 84.20                         & 85.83                         & \multicolumn{1}{c|}{82.90}                         & 82.39                         & 83.47                         & \multicolumn{1}{c|}{80.37}                         & 78.40                         & 85.23                         & 79.93                         \\
\multicolumn{1}{c|}{\multirow{-2}{*}{{\color[HTML]{009900}MPN}}}        & Gap      & -                        & -9.37                         & {\color[HTML]{000000} -11.99} & \multicolumn{1}{c|}{{\color[HTML]{000000} -14.93}} & {\color[HTML]{000000} -14.32} & {\color[HTML]{000000} -12.68} & \multicolumn{1}{c|}{{\color[HTML]{000000} -15.61}} & {\color[HTML]{000000} -16.12} & {\color[HTML]{000000} -15.04} & \multicolumn{1}{c|}{{\color[HTML]{000000} -18.14}} & {\color[HTML]{F56B00} -20.11} & {\color[HTML]{000000} -13.28} & {\color[HTML]{000000} -18.58} \\ \hline
\multicolumn{1}{c|}{}                             & MPJPE            & -                        & 82.70                         & 80.28                         & \multicolumn{1}{c|}{83.93}                         & 82.13                         & 90.73                         & \multicolumn{1}{c|}{84.03}                         & 90.14                         & 84.08                         & \multicolumn{1}{c|}{85.34}                         & 85.09                         & 79.56                         & 83.02                         \\
\multicolumn{1}{c|}{\multirow{-2}{*}{{\color[HTML]{009900}TCMR}}}       & Gap      & -                        & {\color[HTML]{000000} -15.81} & {\color[HTML]{000000} -18.23} & \multicolumn{1}{c|}{{\color[HTML]{000000} -14.58}} & {\color[HTML]{000000} -16.38} & -7.78                         & \multicolumn{1}{c|}{{\color[HTML]{000000} -14.48}} & -8.37                         & {\color[HTML]{000000} -14.43} & \multicolumn{1}{c|}{{\color[HTML]{000000} -13.17}} & {\color[HTML]{000000} -13.43} & {\color[HTML]{000000} -18.95} & {\color[HTML]{000000} -15.49} \\ \hline
\end{tabular}
} 
\end{table*}


\begin{table}[t]
\caption{Quantitative scores of our~\textbf{full scope 3DHR-Co test-time refinement strategy}. We use DynaBOA to perform the bilevel function (\textbf{Line 17}) in our algorithm.~\textbf{MPN} is selected as the intermediate label data extractor. {\color[HTML]{FF0000}Red} mark below defines error gap $<$ -30 mm.} 
\label{table:tab_3DHRCo_MPNpgt_abla}
\resizebox{\linewidth}{!}{%
\begin{tabular}{lccccc}
\hline
\multicolumn{1}{c}{}                                             &                          & \multicolumn{4}{c}{{\color[HTML]{009900}MPN}}                                                       \\ \cline{3-6} 
\multicolumn{1}{c}{\multirow{-2}{*}{Intr.{\color[HTML]{0000ff}-Pert.}\textbackslash{\color[HTML]{009900}-Label}}} & \multirow{-2}{*}{Initial} & MP    & Gap                           & PA    & Gap                           \\ \hline
SPIN                                                             & 98.25 / 60.19            & -     & -                             & -     & -                             \\ \hline
                                                                 & {\color[HTML]{0000ff}$\sigma_1=35$}                     & {\underline{\textbf{63.72}}} & {\color[HTML]{FF0000} -34.53} & 42.11 & {\color[HTML]{000000} -18.08} \\
                                                                 & {\color[HTML]{0000ff}$\sigma_2=50$}                     & 63.87 & {\color[HTML]{FF0000} -34.38} & 42.10 & {\color[HTML]{000000} -18.09} \\
\multirow{-3}{*}{3DHR-Co (SPIN)}                                      & {\color[HTML]{0000ff}$\sigma_3=65$ }                    & 63.89 & {\color[HTML]{FF0000} -34.36} & 42.54 & {\color[HTML]{000000} -17.65} \\ \hline
RSN                                                              & 98.51 / 59.81            & -     & -                             & -     & -                             \\ \hline
                                                                 & {\color[HTML]{0000ff}$\sigma_1=35$}                     & 64.91 & {\color[HTML]{FF0000} -33.60} & 42.11 & {\color[HTML]{000000} -17.07} \\
                                                                 & {\color[HTML]{0000ff}$\sigma_2=50$}                     & 64.77 & {\color[HTML]{FF0000} -33.74} & 42.10 & {\color[HTML]{000000} -17.08} \\
\multirow{-3}{*}{3DHR-Co (RSN)}                                       & {\color[HTML]{0000ff}$\sigma_3=65$}                     & 65.19 & {\color[HTML]{FF0000} -33.32} & 42.54 & {\color[HTML]{000000} -16.64} \\ \hline
\end{tabular}
} 
\end{table}

\begin{table}[t]
\caption{Quantitative scores of our~\textbf{full scope 3DHR-Co test-time refinement strategy}. We use DynaBOA to perform the bilevel function (\textbf{Line 17}) in our algorithm.~\textbf{TCMR} is selected as the intermediate label data extractor. {\color[HTML]{FF0000}Red} mark below defines error gap $<$ -30 mm.}
\label{table:tab_3DHRCo_TCMRpgt_abla}
\resizebox{\linewidth}{!}{%
\begin{tabular}{lccccc}
\hline
\multicolumn{1}{c}{}                                             &                           & \multicolumn{4}{c}{{\color[HTML]{009900}TCMR}}                                                      \\ \cline{3-6} 
\multicolumn{1}{c}{\multirow{-2}{*}{Intr.{\color[HTML]{0000ff}-Pert.}\textbackslash{\color[HTML]{009900}-Label}}} & \multirow{-2}{*}{Initial} & MP    & Gap                           & PA    & Gap                           \\ \hline
SPIN                                                             & 98.25 / 60.19             & -     & -                             & -     & -                             \\ \hline
                                                                 & {\color[HTML]{0000ff}$\sigma_1=35$}                      & 64.21 & {\color[HTML]{FF0000} -34.04} & 41.28 & {\color[HTML]{000000} -18.91} \\
                                                                 & {\color[HTML]{0000ff}$\sigma_2=50$}                      & {\underline{\textbf{63.92}}} & {\color[HTML]{FF0000} -34.34} & 41.14 & {\color[HTML]{000000} -19.05} \\
\multirow{-3}{*}{3DHR-Co (SPIN)}                                      & {\color[HTML]{0000ff}$\sigma_3=65$}                      & 64.13 & {\color[HTML]{FF0000} -34.12} & 41.34 & {\color[HTML]{000000} -18.85} \\ \hline
RSN                                                              & 98.51 / 59.81             & -     & -                             & -     & -                             \\ \hline
                                                                 & {\color[HTML]{0000ff}$\sigma_1=35$}                      & 64.83 & {\color[HTML]{FF0000} -33.68} & 42.05 & {\color[HTML]{000000} -17.13} \\
                                                                 & {\color[HTML]{0000ff}$\sigma_2=50$}                      & 64.82 & {\color[HTML]{FF0000} -33.69} & 41.97 & {\color[HTML]{000000} -17.21} \\
\multirow{-3}{*}{3DHR-Co (RSN)}                                       & {\color[HTML]{0000ff}$\sigma_3=65$}                      & 65.14 & {\color[HTML]{FF0000} -33.37} & 41.90 & {\color[HTML]{000000} -17.28} \\ \hline
\end{tabular}
} 
\end{table}

\begin{figure*}[t]
\begin{center}
        \includegraphics[width=0.96\textwidth]{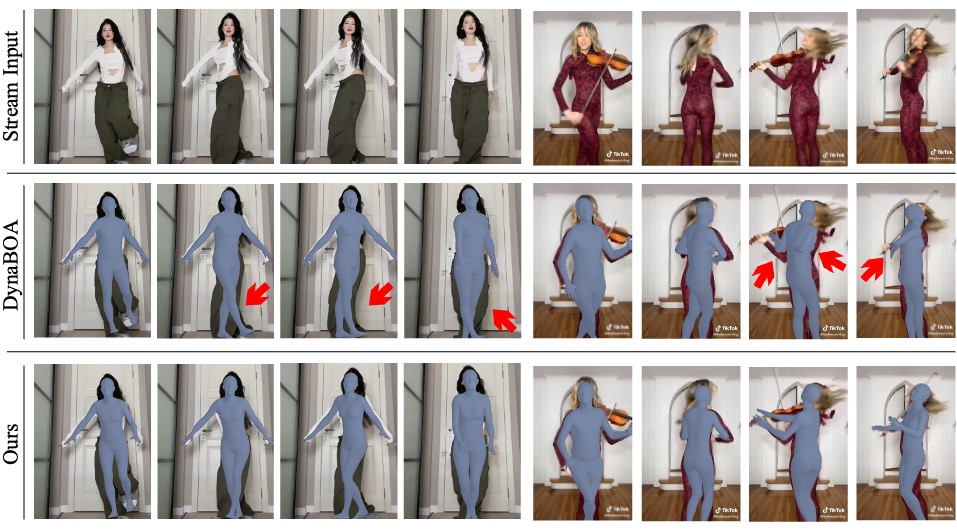} 
\end{center}
\caption{Internet results using our approach that are compared directly recent top performer of 3DHR adaptation method (DynaBOA). Occluded body parts are plausibly estimated using our approach.}
\label{fig:fig06_Internet}
\end{figure*}

\section{Experiment}
\subsection{Experimental Setup}
In this discussion, we provide a comprehensive overview of our implementation method. 
The pseudocode of our algorithm is implemented using the Pytorch framework.
To ensure consistency, various recent and common backbone models are utilized for refinement: SPIN~\cite{Kolotouros_2019_SPIN} and RSN~\cite{Xu_ECCV2020_LRBody}, as they provide the same SMPL parameters and similar Pytorch scripts.
The backbones of these works are directly plugged into our algorithm.
The intermediate label data extractor is performed via temporal $T$ works of MPN~\cite{wei_CVPR2022_mpsnet} and TCMR~\cite{choi_TCMR_CVPR2021}.
The Eq.~\ref{eq:eq_01_loss_preadaptation} is
determined with constant coefficients ($\lambda_1$=10, $\lambda_2$=0.1, $\lambda_3$=1, $\lambda_4$=1) while learned via
ADAM optimizer with a learning rate of 0.00001 during the pre-adaptation stage (from $f_0$ to $f_s$). 
Our regenerated-based bilevel adaptation function (from $f_s$ to $f_a$) follows the recent setting of bilevel function of DynaBOA~\cite{guan_TPAMI2022_dynaboa}.
Our refinement strategy is run on the NVIDIA RTX 3090 GPU, and all experiments are focused on the benchmark of in-the-wild scenario (via 3DPW~\cite{von_3dpw_eccv2018}).

\subsection{Finding the Way for Improvement}
The main challenge of our 3DHR-Co framework is to find reliable data used during test-time learning.
As described in the \textit{Method} Section above, we opt to generate the intermediate label from the temporal-based works while the intermediate perturbed data is obtained from the non-temporal models but with noise addition.
The benefits above can be seen in Figure~\ref{fig:fig03_improvementtosota} where intermediate perturbed data and intermediate label data improve the performance significantly with the MPJPE score of original SPIN \textbf{98.25 mm} boosted to \textbf{91.80 mm} and \textbf{80.25 mm}, respectively, in the 3DPW~\cite{von_3dpw_eccv2018} dataset.
Note that MPJPE is a mean per joint position error metric, which means a lower error score translates to better performance.
Both findings (intermediate-label and -perturbed) are then applied to our pre-adaptation strategy, and it yields a lower error score (up to \textbf{75.76 mm}).
With this capability secured, the pre-adapted version is then used in our full-scope strategy along with the bilevel (Reg. Blv.)~\cite{guan_CVPR2021_bilevelbody} and dynamic bilevel (Reg. Dy. Blv.)~\cite{guan_TPAMI2022_dynaboa} functions.
Their scores are shown in Figure~\ref{fig:fig03_improvementtosota} accordingly.
The recent setting of~\cite{guan_TPAMI2022_dynaboa} is utilized in our bilevel function as it shows best score via full-scope 3DHR-Co framework (refer to right-most result with the score of \textbf{63.72 mm} in Figure~\ref{fig:fig03_improvementtosota}).
Based on this experiment, we further explore the benefit of both pre-adaptation and the full-scope of 3DHR-Co in the following discussions.

\subsection{Pre-adaptation Test-time Refinement Experiment}
The effect of our pre-adaptation strategy is showcased in the following experiment.
This task is performed by working with the intermediate-label and -perturbed data settings.
The intermediate-perturbed data are represented with Gaussian noises varied among $\sigma_1$=35, $\sigma_2$=50, $\sigma_3$=65.
The intermediate label version is represented by the number of sampled data (percentage) taken from the tested in-the-wild dataset.
In this work, each of the epoch sampled 3 video sequence of the 3DPW benchmark set, where each video contains randomized 8 batch of frames, thus, around 24 frames are utilized from the total 35K frames (0.07\%) in 3DPW~\cite{von_3dpw_eccv2018} test set.
We empirically determined the maximum number of ~600 epochs for further pre-adaptation experiments (42\%).

\subsubsection{Working with intermediate-label and intermediate-perturbed data}
We show the quantitative pre-adaptation scores for both SPIN~\cite{Kolotouros_2019_SPIN} and RSN~\cite{Xu_ECCV2020_LRBody} models in Table~\ref{table:tab_SPIN_abla} and Table~\ref{table:tab_RSN_abla} respectively.
The scores are shown using MPJPE metric, varied along the number of perturbations in the horizontal direction and the intermediate label extractor (MPN~\cite{wei_CVPR2022_mpsnet} or TCMR~\cite{choi_TCMR_CVPR2021}) on vertical manner.
As shown in the tables above, the more data to be sampled, the larger improvement is achieved as errors are suppressed in a large gap.
In the case of SPIN model in Table~\ref{table:tab_SPIN_abla}, our pre-adaptation strategy can suppress the error score up to \textbf{-22.49 mm} (orange mark at ep-601), while RSN can be reduced with ~\textbf{-20.11 mm} error gaps (orange mark in Table~\ref{table:tab_RSN_abla}).
This indicates that both of these works have hidden potential without modifications at the architecture level.
The RSN model case, although more focused on the lower-resolution scale 3DHR task, is still capable of doing test-time refinement, as shown in Table~\ref{table:tab_RSN_abla}. 
SPIN's qualitative improvement results from its initial version are demonstrated in Figure~\ref{fig:fig05_SPIN_abla}, while RSN's results are shown in Figure~\ref{fig:fig05_RSN_abla}.
The red arrow marks highlighted the significant pose difference between the initial outputs of the pre-trained model and the refined versions via our strategy.
MPJPE and PA-MPJPE scores are shown directly below the respective image.

We notice that the appreciable improvements are mainly shown in the body parts that are partially occluded.
This phenomenon indicates that our test-time refinement approach in the pre-adaptation stage is already capable of solving the occlusion issue.
In a few cases, however, we observed that larger noise ($\sigma_{3}$) might lead to sub-optimal results (shown in the fourth-row image of Figure~\ref{fig:fig05_SPIN_abla} and Figure~\ref{fig:fig05_RSN_abla}).
Thus, it is recommended to use as low as possible noise value for the 3DPW~\cite{von_3dpw_eccv2018} case. 
This is in line with our finding as quantitatively, both SPIN and RSN demonstrated the best performances via $\sigma_1=35$, as shown in both Table~\ref{table:tab_SPIN_abla} and Table~\ref{table:tab_RSN_abla}, respectively.

\subsection{Full Scope 3DHR-Co Test-time Refinement Experiment}
The next important step in our experiment is to explore the full-scope version of our 3DHR-Co strategy.
Using the pre-adapted weight, our algorithm runs the bilevel function along with the regeneration strategy (\textbf{Line 15-18}.
This approach is proven to give top performance in the current 3DPW~\cite{von_3dpw_eccv2018} benchmark using only classic SPIN backbone (ResNet-50~\cite{He_CVPR2016}).
The quantitative scores are shown in Table~\ref{table:tab_3DHRCo_MPNpgt_abla} and Table~\ref{table:tab_3DHRCo_TCMRpgt_abla}, where MPN and TCMR act as the intermediate label extractor, respectively.
Similar to the discussion above, a large error suppression gap (\textbf{-34.53 mm}) is obtained by 3DHR-Co(SPIN) with the smallest noise configuration (with $\sigma_1$ shown in Table~\ref{table:tab_3DHRCo_MPNpgt_abla}).
This leads the classic SPIN model to achieve superior performance in the current 3DPW~\cite{von_3dpw_eccv2018} benchmark (MPJPE = {\underline{\textbf{63.72 mm}}} as highlighted in Table~\ref{table:tab_3DHRCo_MPNpgt_abla}) with the MPN as an intermediate label data extractor.
Re-plug it with TCMR as an intermediate label data extractor, and the SPIN model is capable of achieving a similar result in the 3DPW benchmark (MPJPE = {\underline{\textbf{63.92 mm}}} as highlighted in Table~\ref{table:tab_3DHRCo_TCMRpgt_abla}).
The works above conclude that the strategy of using 3DHR intermediate data extractors and the pre-adapted classic backbone together is capable of achieving reliable test-time refinement tasks collaboratively in 3DHR works.

\subsection{In-the-wild Real World Experiment}
To further show the validity of our work, 3DHR-Co is demonstrated with real-world internet data cases where no ground truth is available.
We follow the DynaBOA setup that first generates the 2D pose via an off-the-shelf 2D human pose estimator (AlphaPose)~\cite{alphapose_2022,li_2019_crowdpose,fang_2017_rmpe}.
Our method is performed via pre-trained SPIN plugged directly into the 3DHR-Co framework with MPN as the intermediate-label extractor.
As shown in Figure~\ref{fig:fig06_Internet}, the stream inputs that have severe occlusions in the human body are recovered better with our approach compared to DynaBOA~\cite{guan_TPAMI2022_dynaboa}.
The visible errors of~\cite{guan_TPAMI2022_dynaboa} (red-arrows in Figure~\ref{fig:fig06_Internet}) are the issue of human pose ambiguity problem where some occluded paired-body parts are often switched.
Our 3DHR-Co approach solved this issue while showing considerable temporal results from the input stream (refer to \textit{supplementary files}).

\subsection{Execution Time}
Our approach came with the extra advantage as it runs with small computational time.
Our pre-adaptation approach took approximately 0.15 seconds for a batch of data (each batch contains 8 frames with the size of $3\times224\times224$ dimensions). 
For an epoch that sampled $3\times8$ frames takes around 3 seconds for each pre-adaptation step.
As shown in Table~\ref{table:tab_SPIN_abla}, running approximately 400 epochs in the pre-adaptation step takes only less than or around half-hour to boost the performance of pre-trained SPIN~\cite{Kolotouros_2019_SPIN} up to \textbf{-21 mm} error suppression on the whole 3DPW~\cite{von_3dpw_eccv2018} test set.
Such an approach is favorable compared to re-training a new 3DHR model that takes hours or days to learn.
The regeneration-based bilevel steps share similar timing with the original source~\cite{guan_TPAMI2022_dynaboa} that take approximately 1 second for adapting a batch of data (1 batch runs  1 frame).

\subsection{Limitations}
While the experiments above show remarkable performances in boosting the 3DHR models, our current development is still limited in terms of boosting test data with the non in-the-wild characteristics.
The performance of the full-scope 3DHR-Co framework depends on the intermediate data extraction strategy applied during the pre-adaptation stage.
In the pre-adaptation stage, when MPI~\cite{andriluka_mpii_cvpr2014} and Human3.6M~\cite{ionescu_TPAMI14_human36m} are tested directly, our refinement performances are sub-optimal as it only improved with only around \textbf{-2 mm} in both cases.
We presume this phenomenon happened due to their dataset characteristic that mostly contains homogeneous scenes from in-studio environments.
Specifically, their scenes are equipped with static backgrounds and similar color information shared among videos and frames.
This condition is well fitted on both networks during the training stage (via~\cite{ionescu_TPAMI14_human36m} training set), and thus, adding noises to such scenes gives minimum effect during test-time refinement.
Future studies should consider the constraint above to optimally perform 3DHR refinement in various dataset domains.

\section{Conclusion}
In this work, we propose a collaborative-based test-time refinement framework (termed 3DHR-Co) specialized for boosting the 3DHR task.
Running 3DHR in an in-the-wild scenario poses an important challenge, as it is hard to obtain 3D human pose labels for training the 3DHR model.
Such a case subsequently pushes the recent 3DHR works and their refinement approaches to learning with limited supervision.
The 3DHR-Co is proposed to answer this challenge by allowing various common 3DHR models to be boosted directly in test time.
This is achieved by employing 2 technical strategies: (i) finding the way of extracting reliable intermediate data during test-time to be used during test-time learning and (ii) the refinement framework itself that leverages the collaborative strategy.
In specific, the collaborative approach involves the process of transferring the knowledge of one 3DHR model to another.
Our experimental results showed that even with the common classic 3DHR models, our framework obtained a remarkable performance boost, achieving top performance results and thereby revealing their true potential in solving the in-the-wild scenario.
The explorations above are also provided with thorough discussions to help find the best test-time refined 3DHR outcomes using our method.
We also describe our work limitations for future improvement of 3DHR refinement studies.


\begin{thebibliography}{10}\itemsep=-1pt

\bibitem{alldieck_photorealistic_cloth_cvpr2022}
Thiemo Alldieck, Mihai Zanfir, and Cristian Sminchisescu.
\newblock Photorealistic monocular 3d reconstruction of humans wearing clothing.
\newblock In {\em Computer Vision and Pattern Recognition}, pages 1506--1515, 2022.

\bibitem{andriluka_mpii_cvpr2014}
Mykhaylo Andriluka, Leonid Pishchulin, Peter Gehler, and Bernt Schiele.
\newblock 2d human pose estimation: New benchmark and state of the art analysis.
\newblock In {\em Computer Vision and Pattern Recognition}, pages 3686--3693, 2014.

\bibitem{choi_TCMR_CVPR2021}
Hongsuk Choi, Gyeongsik Moon, Ju~Yong Chang, and Kyoung~Mu Lee.
\newblock Beyond static features for temporally consistent 3d human pose and shape from a video.
\newblock In {\em Computer Vision and Pattern Recognition}, pages 1964--1973, 2021.

\bibitem{choi2020pose2mesh}
Hongsuk Choi, Gyeongsik Moon, and Kyoung~Mu Lee.
\newblock Pose2mesh: Graph convolutional network for 3d human pose and mesh recovery from a 2d human pose.
\newblock In {\em European Conference on Computer Vision}, pages 769--787, 2020.

\bibitem{ehret_frame2frame_cvpr2019}
Thibaud Ehret, Axel Davy, Jean-Michel Morel, Gabriele Facciolo, and Pablo Arias.
\newblock Model-blind video denoising via frame-to-frame training.
\newblock In {\em Computer Vision and Pattern Recognition}, pages 11369--11378, 2019.

\bibitem{alphapose_2022}
Hao-Shu Fang, Jiefeng Li, Hongyang Tang, Chao Xu, Haoyi Zhu, Yuliang Xiu, Yong-Lu Li, and Cewu Lu.
\newblock Alphapose: Whole-body regional multi-person pose estimation and tracking in real-time.
\newblock {\em IEEE Transactions on Pattern Analysis and Machine Intelligence}, 2022.

\bibitem{fang_2017_rmpe}
Hao-Shu Fang, Shuqin Xie, Yu-Wing Tai, and Cewu Lu.
\newblock {RMPE}: Regional multi-person pose estimation.
\newblock In {\em International Conference on Computer Vision}, 2017.

\bibitem{fang2021reconstructing}
Qi Fang, Qing Shuai, Junting Dong, Hujun Bao, and Xiaowei Zhou.
\newblock Reconstructing 3d human pose by watching humans in the mirror.
\newblock In {\em Computer Vision and Pattern Recognition}, 2021.

\bibitem{guan_TPAMI2022_dynaboa}
Shanyan Guan, Jingwei Xu, Michelle~Zhang He, Yunbo Wang, Bingbing Ni, and Xiaokang Yang.
\newblock Out-of-domain human mesh reconstruction via dynamic bilevel online adaptation.
\newblock {\em IEEE Transactions on Pattern Analysis and Machine Intelligence}, 45(4):5070--5086, 2022.

\bibitem{guan_CVPR2021_bilevelbody}
Shanyan Guan, Jingwei Xu, Yunbo Wang, Bingbing Ni, and Xiaokang Yang.
\newblock Bilevel online adaptation for out-of-domain human mesh reconstruction.
\newblock In {\em Computer Vision and Pattern Recognition}, pages 10472--10481, 2021.

\bibitem{He_CVPR2016}
Kaiming He, Xiangyu Zhang, Shaoqing Ren, and Jian Sun.
\newblock Deep residual learning for image recognition.
\newblock In {\em Computer Vision and Pattern Recognition}, pages 770--778, 2016.

\bibitem{he_archplus_iccv2021}
Tong He, Yuanlu Xu, Shunsuke Saito, Stefano Soatto, and Tony Tung.
\newblock Arch++: Animation-ready clothed human reconstruction revisited.
\newblock In {\em International Conference on Computer Vision}, pages 11046--11056, 2021.

\bibitem{ionescu_TPAMI14_human36m}
Catalin Ionescu, Dragos Papava, Vlad Olaru, and Cristian Sminchisescu.
\newblock Human3.6{M}: Large scale datasets and predictive methods for 3{D} human sensing in natural environments.
\newblock {\em IEEE Trans. on Pattern Analysis and Machine Intelligence}, 36(7):1325--1339, jul 2014.

\bibitem{iqbal2020weakly}
Umar Iqbal, Pavlo Molchanov, and Jan Kautz.
\newblock Weakly-supervised 3d human pose learning via multi-view images in the wild.
\newblock In {\em Computer Vision and Pattern Recognition}, 2020.

\bibitem{hanbyul_expressivemodel_CVPR2018}
Hanbyul Joo, Tomas Simon, and Yaser Sheikh.
\newblock Total capture: A 3{D} deformation model for tracking faces, hands, and bodies.
\newblock In {\em Computer Vision and Pattern Recognition}, pages 8320--8329, 2018.

\bibitem{kanazawa_CVPR18}
Angjoo Kanazawa, Michael~J. Black, David~W. Jacobs, and Jitendra Malik.
\newblock End-to-end recovery of human shape and pose.
\newblock In {\em Computer Vision and Pattern Recognition}, pages 7122--7131, 2018.

\bibitem{kanazawa_CVPR19}
Angjoo Kanazawa, Jason~Y. Zhang, Panna Felsen, and Jitendra Malik.
\newblock Learning 3{D} human dynamics from video.
\newblock In {\em Computer Vision and Pattern Recognition}, pages 5614--5623, 2019.

\bibitem{kocabas_vibe_CVPR2020}
Muhammed Kocabas, Nikos Athanasiou, and Michael~J Black.
\newblock Vibe: Video inference for human body pose and shape estimation.
\newblock In {\em Computer Vision and Pattern Recognition}, pages 5253--5263, 2020.

\bibitem{kocabas_pare_ICCV2021}
Muhammed Kocabas, Chun-Hao~P Huang, Otmar Hilliges, and Michael~J Black.
\newblock Pare: Part attention regressor for 3d human body estimation.
\newblock In {\em International Conference on Computer Vision}, pages 11127--11137, 2021.

\bibitem{Kolotouros_2019_SPIN}
Nikos Kolotouros, Georgios Pavlakos, Michael~J Black, and Kostas Daniilidis.
\newblock Learning to reconstruct 3d human pose and shape via model-fitting in the loop.
\newblock In {\em International Conference on Computer Vision}, pages 2252--2261, 2019.

\bibitem{lee_restore_pseudo_CVPR2021}
Seunghwan Lee, Donghyeon Cho, Jiwon Kim, and Tae~Hyun Kim.
\newblock Restore from restored: Video restoration with pseudo clean video.
\newblock In {\em Computer Vision and Pattern Recognition}, pages 3537--3546, 2021.

\bibitem{Lee_DynaVidSR_WACV21}
Suyoung Lee, Myungsub Choi, and Kyoung~Mu Lee.
\newblock Dynavsr: Dynamic adaptive blind video super-resolution.
\newblock In {\em Winter Conference on Applications of Computer Vision}, pages 2093--2102, 2021.

\bibitem{li_2019_crowdpose}
Jiefeng Li, Can Wang, Hao Zhu, Yihuan Mao, Hao-Shu Fang, and Cewu Lu.
\newblock Crowdpose: Efficient crowded scenes pose estimation and a new benchmark.
\newblock In {\em Computer Vision and Pattern Recognition}, pages 10863--10872, 2019.

\bibitem{lin2014mscoco}
Tsung-Yi Lin, Michael Maire, Serge Belongie, James Hays, Pietro Perona, Deva Ramanan, Piotr Doll{\'a}r, and C~Lawrence Zitnick.
\newblock Microsoft coco: Common objects in context.
\newblock In {\em European Conference on Computer Vision}, 2014.

\bibitem{loper_ACM15}
Matthew Loper, Naureen Mahmood, Javier Romero, Gerard Pons{-}Moll, and Michael~J. Black.
\newblock {SMPL:} a skinned multi-person linear model.
\newblock {\em {ACM} {T}rans. on {G}raphics}, 34(6):248:1--248:16, 2015.

\bibitem{luo_meva_ACCV2020}
Zhengyi Luo, S~Alireza Golestaneh, and Kris~M Kitani.
\newblock 3d human motion estimation via motion compression and refinement.
\newblock In {\em Asian Conference on Computer Vision}, 2020.

\bibitem{mehta2018single}
Dushyant Mehta, Oleksandr Sotnychenko, Franziska Mueller, Weipeng Xu, Srinath Sridhar, Gerard Pons-Moll, and Christian Theobalt.
\newblock Single-shot multi-person 3d pose estimation from monocular rgb.
\newblock In {\em 3DV}, 2018.

\bibitem{Moon_2020_I2L}
Gyeongsik Moon and Kyoung~Mu Lee.
\newblock I2l-meshnet: Image-to-lixel prediction network for accurate 3d human pose and mesh estimation from a single rgb image.
\newblock In {\em European Conference on Computer Vision}, pages 752--768. Springer, 2020.

\bibitem{KimFast_ECCV2020}
Seobin Park, Jinsu Yoo, Donghyeon Cho, Jiwon Kim, and Tae~Hyun Kim.
\newblock Fast adaptation to super-resolution networks via meta-learning.
\newblock In {\em European Conference on Computer Vision}. Springer, 2020.

\bibitem{pavlakos_smplx_cvpr2019}
Georgios Pavlakos, Vasileios Choutas, Nima Ghorbani, Timo Bolkart, Ahmed~AA Osman, Dimitrios Tzionas, and Michael~J Black.
\newblock Expressive body capture: 3d hands, face, and body from a single image.
\newblock In {\em Computer Vision and Pattern Recognition}, pages 10975--10985, 2019.

\bibitem{rhodin2018unsupervised}
Helge Rhodin, Mathieu Salzmann, and Pascal Fua.
\newblock Unsupervised geometry-aware representation for 3d human pose estimation.
\newblock In {\em European Conference on Computer Vision}, pages 750--767, 2018.

\bibitem{rhodin2018learning}
Helge Rhodin, J{\"o}rg Sp{\"o}rri, Isinsu Katircioglu, Victor Constantin, Fr{\'e}d{\'e}ric Meyer, Erich M{\"u}ller, Mathieu Salzmann, and Pascal Fua.
\newblock Learning monocular 3d human pose estimation from multi-view images.
\newblock In {\em Computer Vision and Pattern Recognition}, 2018.

\bibitem{romero2022embodied_siggraph2017}
Javier Romero, Dimitrios Tzionas, and Michael~J Black.
\newblock Embodied hands: Modeling and capturing hands and bodies together.
\newblock {\em SIGGRAPH}, 2017.

\bibitem{Shunsuke_2019_PIFu}
Shunsuke Saito, Zeng Huang, Ryota Natsume, Shigeo Morishima, Angjoo Kanazawa, and Hao Li.
\newblock Pifu: Pixel-aligned implicit function for high-resolution clothed human digitization.
\newblock In {\em International Conference on Computer Vision}, pages 2304--2314, 2019.

\bibitem{saito_pifuhd_cvpr2020}
Shunsuke Saito, Tomas Simon, Jason Saragih, and Hanbyul Joo.
\newblock Pifuhd: Multi-level pixel-aligned implicit function for high-resolution 3d human digitization.
\newblock In {\em Computer Vision and Pattern Recognition}, pages 84--93, 2020.

\bibitem{shocherzero_cvpr2018}
Assaf Shocher, Nadav Cohen, and Michal Irani.
\newblock “zero-shot” super-resolution using deep internal learning.
\newblock In {\em Computer Vision and Pattern Recognition}, pages 3118--3126, 2018.

\bibitem{SohMeta_CVPR2020}
Jae~Woong Soh, Sunwoo Cho, and Nam~Ik Cho.
\newblock Meta-transfer learning for zero-shot super-resolution.
\newblock In {\em Computer Vision and Pattern Recognition}, pages 3516--3525, 2020.

\bibitem{sun_catastroph_forget_ICML2020}
Yu Sun, Xiaolong Wang, Zhuang Liu, John Miller, Alexei Efros, and Moritz Hardt.
\newblock Test-time training with self-supervision for generalization under distribution shifts.
\newblock In {\em International Conference on Machine Learning}, pages 9229--9248, 2020.

\bibitem{von_3dpw_eccv2018}
Timo von Marcard, Roberto Henschel, Michael~J Black, Bodo Rosenhahn, and Gerard Pons-Moll.
\newblock Recovering accurate 3d human pose in the wild using imus and a moving camera.
\newblock In {\em European Conference on Computer Vision}, pages 601--617. Springer, 2018.

\bibitem{wandt2019repnet}
Bastian Wandt and Bodo Rosenhahn.
\newblock Repnet: Weakly supervised training of an adversarial reprojection network for 3d human pose estimation.
\newblock In {\em Computer Vision and Pattern Recognition}, 2019.

\bibitem{wandt2021canonpose}
Bastian Wandt, Marco Rudolph, Petrissa Zell, Helge Rhodin, and Bodo Rosenhahn.
\newblock Canonpose: Self-supervised monocular 3d human pose estimation in the wild.
\newblock In {\em Computer Vision and Pattern Recognition}, 2021.

\bibitem{wei_CVPR2022_mpsnet}
Wen-Li Wei, Jen-Chun Lin, Tyng-Luh Liu, and Hong-Yuan~Mark Liao.
\newblock Capturing humans in motion: temporal-attentive 3d human pose and shape estimation from monocular video.
\newblock In {\em Computer Vision and Pattern Recognition}, pages 13211--13220, 2022.

\bibitem{xu_cvpr2020_ghuml}
Hongyi Xu, Eduard~Gabriel Bazavan, Andrei Zanfir, William~T Freeman, Rahul Sukthankar, and Cristian Sminchisescu.
\newblock Ghum \& ghuml: Generative 3d human shape and articulated pose models.
\newblock In {\em Computer Vision and Pattern Recognition}, pages 6184--6193, 2020.

\bibitem{Xu_ECCV2020_LRBody}
Xiangyu Xu, Hao Chen, Francesc Moreno-Noguer, Laszlo~A. Jeni, and Fernando~De la Torre.
\newblock 3d human shape and pose from a single low-resolution image with self-supervised learning.
\newblock In {\em European Conference on Computer Vision}. Springer, 2020.

\bibitem{yao2019monet}
Yuan Yao, Yasamin Jafarian, and Hyun~Soo Park.
\newblock Monet: Multiview semi-supervised keypoint detection via epipolar divergence.
\newblock In {\em International Conference on Computer Vision}, 2019.

\end{thebibliography}


{\small




}

\end{document}